# Generating meta-learning tasks to evolve parametric loss for classification learning

Author: Zhaoyang Hai, Xiabi Liu*, Yuchen Ren, Nouman Q. Soomro

**Abstract**

The field of meta-learning has seen a dramatic rise in interest in recent years. In existing meta-learning approaches, learning tasks for training meta-models are usually collected from public datasets, which brings the difficulty of obtaining a sufficient number of meta-learning tasks with a large amount of training data. In this paper, we propose a meta-learning approach based on randomly generated meta-learning tasks to obtain a parametric loss for classification learning based on big data. The loss is represented by a deep neural network, called meta-loss network (MLN). To train the MLN, we construct a large number of classification learning tasks through randomly generating training data, validation data, and corresponding ground-truth linear classifier. Our approach has two advantages. First, sufficient meta-learning tasks with large number of training data can be obtained easily. Second, the ground-truth classifier is given, so that the difference between the learned classifier and the ground-truth model can be measured to reflect the performance of MLN more precisely than validation accuracy. Based on this difference, we apply the evolutionary strategy algorithm to find out the optimal MLN. The resultant MLN not only leads to satisfactory learning effects on generated linear classifier learning tasks for testing, but also behaves very well on generated nonlinear classifier learning tasks and various public classification tasks. Our MLN stably surpass cross-entropy (CE) and mean square error (MSE) in testing accuracy and generalization ability. These results illustrate the possibility of achieving satisfactory meta-learning effects using generated learning tasks.

## 1 Introduction

In the last couple of years, meta-learning has received extensive attention for its ability to automatically optimize machine learning methods and has been applied to various domains such as supervised learning and reinforcement learning. Since meta-learning tasks are constructed based on public datasets at present, it is difficult to collect enough learning tasks with a lot of training data for training meta-models. Although this problem has not yet seriously affected the meta-learning methods for few-shot learning [1-3], it hinders the development of meta-learning for supervised learning on big data, which is also needed to be improved.

In this paper, we propose a meta-learning approach based on randomly generated meta-learning tasks to obtain a parametric loss, which is represented by a deep neural network and called meta-loss network (MLN). Our approach is illustrated in Fig.1 which consisted of four steps. First, many classifier learning tasks are generated randomly, each of which is composed of a set of training data, a set of validation data, and a ground-truth linear classifier (single-layer perceptron). Second, the meta-training is conducted with a process composed of the inner loop and the outer loop. In the inner loop, the $MLN$ is used as the loss function to optimize a classifier iteratively. In the outer loop, the $MLN$ is evaluated by the difference between the learned classifier and the underline ground-truth classifier and optimized by Evolutionary Strategy (ES). To evaluate the proposed meta-learning approach, the resultant $MLN$ is tested by randomly generated meta-testing tasks (single-layer classifier and multi-layer classifier), as well as various real tasks on public datasets with popular deep networks.

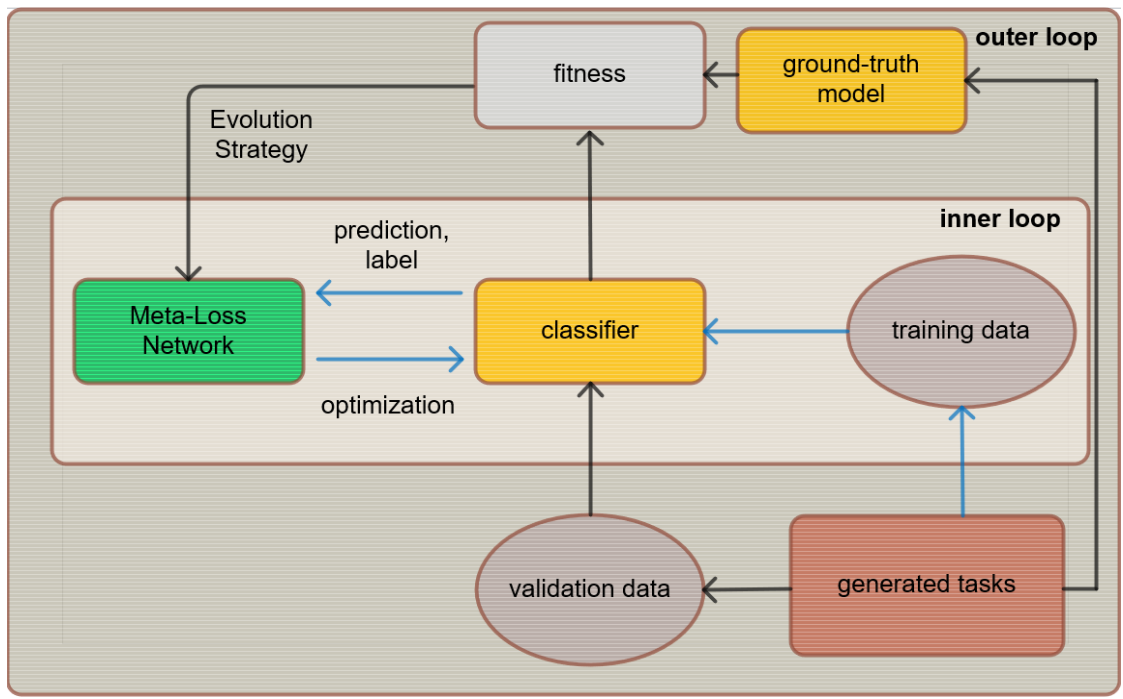

Figure1. The architecture of the proposed meta-learning method.

Our main contributions are summarized as follows.

1) We propose a meta-learning method based on randomly generated learning tasks and show the possibility that randomly generated tasks can provide appropriate meta-knowledge, which helps to overcome the difficulty of obtaining large amounts of supervised learning tasks on big data. To our best knowledge, this is the first time of using randomly generated learning tasks for meta-learning.

2) We obtain a satisfactory parametric loss function for classification learning on big data, called MLN, by using the proposed meta-learning approach. Benefitting from generated learning tasks of linear classifier, we design a novel objective for training the MLN, targeting at minimizing the difference between the learned classifier and the ground-truth classifier on validation data. This goal implies there is only one choice about the optimal result in linear classifier learning, which gives a strict constraint on the searching of optimal MLN. Thus, the better and more generalized meta-knowledge can be found and represented in the learned MLN. It brings an amazing generalization performance. Even if trained on simplest linear classification tasks, our optimal MLN outperforms the widely-used cross-entropy (CE) and mean square error (MSE) on much more complicated classification learning tasks.

**2 Related work**

In this section, we first introduce the main meta-learning methods for supervised learning and then introduce the meta-learning methods for automatic acquisition of objective functions.

2.1 Meta-learning for few-shot learning

In the field of supervised learning, existing meta-learning methods are mainly developed for few-shot learning. These can be divided into two categories: optimization-based methods and metric-based methods.

The optimization-based methods are also known as learning to learn, which tackles the few-shot classification problems by effectively optimizing model parameters to new tasks. Finn et al. [1] proposed a model-agnostic algorithm called MAML, for fast adaptation to any novel task through a few gradient steps. LEE et al. [4] set a meta-learner for each layer of the task network so that each layer completes gradient descent optimization under the guidance of a specific meta-learner. Grant et al. [5] redefined MAML and combined it with the probabilistic inference method in the hierarchical Bayesian model. By using the extensible gradient descent process for posterior inference, it is suitable for complex function approximation problems. Andrychowicz et al. [2] and Ravi et al. [6] used a recurrent neural network to replace traditional optimization algorithms in deep learning. Munkhdalai et al. [7] improved on the model by combining fast weights (predicted per task by a network) and slow weights (trained with REINFORCE across tasks) to access memory. Rajasegaran et al. [8] proposed a task-agnostic meta-learning algorithm to minimize catastrophic forgetting. It learns a task-agnostic model that predicts the tasks automatically and subsequently adapts to the predicted tasks. Finally, it mitigates the data imbalance problem by tuning the task-specific parameters separately for each task. Bronskill et al. [9] proposed a task normalization method for meta-learning. They calculate the mean and variance from all support sets and extend them to target sets after weighted summations, whose weights are learnable.

The family of metric-based methods aims to address few-shot classification by learning to compare. Vinyals et al. [10] mapped the support set via an attention mechanism to a function and then classified the query sample by a weighted nearest-neighbor classifier in an embedding space. Snell et al. [11] followed a similar idea that learns a metric-based prediction rule over embeddings. The prototype of each category is represented by computing distances to the nearest category mean. Sung et al. [12] introduced an additional parameterized CNN-based 'relation module' for learnable metric comparison.

2.2 Meta-learning for objectives

In meta-learning methods for automatic acquisition of objective functions, Bechtle et al. [13] proposed a differentiable framework of meta-learning to optimize a parametric loss and applied it to supervised learning and reinforcement learning. A feedforward neural network with 2 hidden layers is used as the loss function and trained for binary classification tasks. They used the MSE for predictions and labels of training data to optimize the meta-model; Houthooft et al. [14] represented the loss function using the network consisting of temporal convolutional layers and dense layers and optimized it through ES algorithm. They constructed a meta-learning framework including inner loop and outer loop. A reinforcement learning process is simulated in the inner loop, where the meta-model is used to replace the Bellman equation. In the outer loop, the parameter of the meta model is updated according to the rewards from inner loops; The goal of Wu et al. [15] is a teacher network that can dynamically generate different loss functions for a machine learning model (student). The teacher network takes the state of the student network at each time step (such as classification accuracy and iteration times) as input, and output a loss function to optimize the student network; In addition to the parameterized loss function, Gonzalez et al. [3] built loss functions hierarchically from a set of operators and leaf nodes. These functions are repeatedly recombined and mutated to find an optimal structure, and then a covariance-matrix adaptation evolutionary strategy (CMA-ES) is used to find optimal coefficients.

3 The proposed Method

3.1 Expressing the loss function by neural network

In our method, the *MLN* is introduced to represent the loss function for training classifiers. We considered training MLN as a binary classification loss, which can be extended to multiple classification tasks by the 1-versus-1 strategy. Because of the powerful representation capability of deep networks, we consider using multi-layer fully connection networks as a meta-model. After careful experiments, we end up with a 6-layer fully connected network as the MLN. The final architecture of our MLN is shown in Fig. 2, where $a_k^{[n]}$ means the k-th neuron in the n-th layer. The input of this network is the prediction from the classifier and the corresponding one-hot label, and the output is the corresponding loss. The numbers of neurons in each hidden layer are 32, 64, 128, 256, and 512, respectively. The Parametric Rectified Linear Unit (PReLU) is used as the activation function of the hidden layers, and the SoftPuls [16] is used for the output layer.

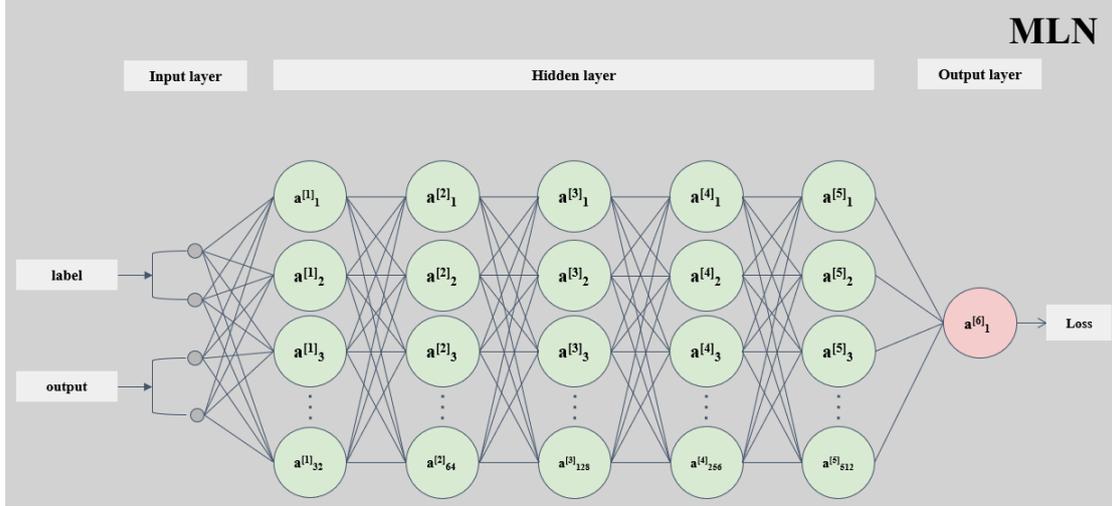

Figure 2. The architecture of our $MLN$.

### 3.2 Constructing the meta-learning tasks

In this section, the construction of meta-learning tasks for training and testing MLN is discussed. At first, we collect 50 normal distributions whose mean and standard deviation are sampled from [0, 5] randomly. When generating each data point, we randomly sampled a normal distribution from 50 distributions and sampled 5 values as features of the data point from this distribution. In this way, we generated two sets of 5-dimensional data points. Then, the first set is used to construct meta-training tasks, and the second is used to construct meta-testing tasks. The data set for meta-training or meta-testing is further divided into a training dataset and a validation dataset. Accordingly, each binary classifier learning task for meta-training or meta-testing is generated as follows.

First, a subset of training data is sampled from the training dataset and a subset of validation data is sampled from the validation dataset.

Second, we randomly initialize a set of ground-truth models with the Xavier [17]. For achieving the better generalization of learned meta-knowledge, we choose the single-layer perceptron for training the MLN. Let $\theta$ be the parameters of a single-layer perceptron $C$, $C_\theta(x)$ be the prediction of $C$ for the data point $x$, then the gradient for optimizing single-layer perception based on the MLN is shown in Eq.1.

$$\frac{\partial MLN}{\partial \theta} = \frac{\partial MLN}{\partial C_\theta(x)} \frac{\partial C_\theta(x)}{\partial \theta} = \frac{\partial MLN}{\partial C_\theta(x)} x \qquad (1)$$

Eq. 1 shows that two factors are affecting the optimization effect: $\frac{\partial MLN}{\partial C_\theta(x)}$ and $x$, both of which are independent of the classifier. So, there is no information about the classifier structure in the use of MLN in meta-learning tasks. It makes the learning of MLN unaffected by special classifier structures and guarantees good generalization of learned MLN.

Third, each ground-truth model is used to divide the training dataset into two categories. As a result, a classifier learning task is constructed with a set of training data, a set of validation data, and a ground-truth classifier. To avoid the side-effect of imbalanced learning to MLN training, class-imbalance tasks are not involved.

We summarize the above process of generating meta-learning tasks in Algorithm 1.

---

**Algorithm 1**: Generating classifier learning tasks for meta-training and meta-testing

**initialize:** the set of ground-truth linear classifiers $set(C_{\theta_{groundTruth}})$, meta-training dataset ($sample_{training,0,...,K}$), meta-testing dataset ($sample_{testing,0,...,K}$)

**Divide** $sample_{training,0,...,K}$ **into** training dataset $\mathfrak{D}_{training\ data}$ and validation dataset $\mathfrak{D}_{validation\ data}$

**Divide** $sample_{testing,0,...,K}$ **into** training dataset $\mathfrak{T}_{training\ data}$ and validation dataset $\mathfrak{T}_{validation\ data}$

**if** sampling for meta training:
  **random sampling** *training data* from $\mathfrak{D}_{training\ data}$
  **random sampling** *validation data* from $\mathfrak{D}_{validation\ data}$

**if** sampling for meta testing:
  **random sampling** *training data* from $\mathfrak{T}_{training\ data}$
  **random sampling** *validation data* from $\mathfrak{T}_{validation\ data}$

**random sampling** a ground truth classifier $C_{\theta^*}$ from $set(C_{\theta_{groundTruth}})$

label_training = $C_{\theta^*}(training\ data)$

label_validation = $C_{\theta^*}(validation\ data)$

**Output:** (*training data*, label_training), (*validation data*, label_validation), $C_{\theta^*}$

---

The advantages of the above-proposed method for constructing meta-learning tasks include 1) compared with using public classification tasks, it is much easier to construct a lot of generated tasks with huge data points; 2) the ground-truth classifier is known in each learning task, through which we can evaluate the performance of the MLN more flexibly and accurately than the case with unknown ground-truth classifier.

### 3.3 The Meta-Learning Objective

After optimizing the classifier with the MLN, the mean squared error (MSE) between the outputs of the optimized classifier and that of the ground-truth classifier on validation data is used as meta-loss to reflect the performance of MLN. The corresponding computation for optimizing MLN is expressed by Eq. 2, where $m \sim p(M)$ denotes a learning task constructed using the method in Section 3.2, C(validation data) is the prediction on validation data by the optimized classifier, and $C_{\theta^*}(\text{validation data})$ is the prediction of a ground-truth classifier for the validation data.

$$\phi^* = \underset{\phi}{\arg\max}\, \mathbb{E}_{m \sim p(M)} \mathbb{E}_{val,label \sim m, MLN}[-\text{MSE}\,(C(\text{validation data}), C_{\theta^*}(\text{validation data}))] \tag{2}$$

This goal implies there is only one choice about the optimal result in linear classifier learning, which gives a strict constraint on the searching of optimal MLN. Thus, the better and more generalized meta-knowledge can be found and represented in the learned MLN. We will see that it brings an amazing generalization performance in our experiments.

### 3.4 Evolving of MLN

To obtain the optimal MLN according to the criterion described in the last section, we adopt the ES algorithm in the outer loop of our framework. Particularly, We apply the classical $(\mu + \lambda)\,ES$ optimization [18] to our problem. The corresponding MLN optimization algorithm, i.e., the whole process of MLN meta-learning, is given in Algorithm 2. As shown there, At first, we initialize $\mu$ MLN and corresponding mutation strength $\sigma$. Then, at the start of each outer loop, $\mu$ MLN will produce $\lambda$ children according to $(\mu + \lambda)\,ES$. Given a loss function $MLN_i, i \in \{1, \ldots, \mu + \lambda\}$, we perform a classifier learning task in the inner loop, which is randomly sampled from the meta-training set. At the end of the inner loop, we evaluate the MLN by using Eq. 1 to get its fitness which is used to select $\mu$ MLN in the outer loop.

| Algorithm 2: Meta-learning of MLN with evolutionary strategy |
|---|
| **Hyper-parameter** number of parents/survivors μ, number of children λ, number of inner loop N, number of outer loop E |
| **Initialize** $MLN_{1,\ldots,\mu}$, mutation strength $\sigma_{1,\ldots,\mu}$ |
| **[Outer Loop]** |
| **for** episode=(1, … , E) **do** |
|     **Producing** λ children according to ES and **mixing** them with parents: $MLN_{1,\ldots,\mu+\lambda}, \sigma_{1,\ldots,\mu+\lambda}$ |
|     **[Inner Loop]** |
|     **for** worker=(1,…, μ + λ) **do** |
|         **Initialize** classifier C |
|         Sampling a meta-training task according to **Algorithm 1**: (training data, label_training), (validation data, label_validation), $C_{\theta^*}$ |
|         **for** step=(1,…,N) **do** |
|             prediction = C(training data) |
|             loss = $MLN_{\text{worker}}(prediction, label\_training)$ |
|             Optimize C to minimize the above loss |

>        **End for**
>        performance$_{\text{worker}}$ = −MSE (C(validation data), $C_{\theta^*}$(validation data))
>        recording performance$_{\text{worker}}$ as the **fitness** of $MLN_{\text{worker}}$
>    **End for**
>    $MLN_{1,\ldots,\mu}$, $\sigma_{1,\ldots,\mu}$ = **selecting** μ $MLN$ and $\sigma$ according to **fitness**
> **End for**
> **Output:**   the optimal $MLN_\phi$

### 3.3 Using the optimized MLN for learning classifiers

The optimized MLN can be applied to classification tasks (our simulated meta-testing tasks or actual tasks) by using it to replace traditional loss functions such as CE and MSE. Since the MLN is trained as a binary-classification loss, when facing a multi-classification task, we convert the task into binary classification tasks with the 1-vs-1 strategy, as shown in Eq.3.

$$Loss = \sum_i MLN([true\_probability, false\_probability_i, 1, 0]) \quad (3)$$

which means we consider all the binary losses between the correct category and each of the wrong for each sample. The corresponding MLN-based classifier learning procedure is given in Algorithm 3.

> Algorithm 3: $MLN$ based classification learning
> ---
> **input** training data x, label y
> **hyper-parameter** number of categories n, batch size K, training step S
> **Initialize** classifier C
> **for** step=(1, …, S) **do**
>     pred = C(train_x)
>     for j=(1, …, K)
>         getting the prediction of correct category in pred: $true\_pre$
>         getting the predictions of wrong category in pred: $(false\_pre_1, \ldots, false\_pre_{n-1})$
>     $loss = \frac{1}{K(n-1)} \sum_{j=1}^{K} \sum_{i=1}^{n-1} MLN([true\_pre_j, false\_pre_{ji}, 1, 0])$
>     Optimize C to minimize the above loss
> **End for**
> **Output:**   optimal C

## 4 Experiments

### 4.1 Experimental Setup

To evaluate the effectiveness of the proposed approach, we use it to train an MLN and test the optimized MLN on various tasks of deep classifier training based on gradient descent. The corresponding computation of optimizing the classifier is

$$\theta = \theta - optimizer(\nabla_\theta(MLN_\phi(\mathcal{C}_\theta(x), y))), \quad (4)$$

where $\theta$ is the parameters of classifier, $optimizer$ is the optimization algorithm in classification learning, and $\nabla_\theta(MLN_\phi(\mathcal{C}_\theta(x), y))$ is the gradient of the MLN loss to parameters of the classifier.

In the meta-training phase, the randomly generated training dataset and validation dataset contain 500,000 samples, respectively. Each learning task for training MLN is constructed randomly by sampling 50,000 training samples, 10,000 validation samples, and a ground-truth model (refer to Algorithm 1).

In the meta-testing phase, generated learning tasks and real learning tasks are performed. In generated tasks, we consider using single-layer perception to prove that MLN learns the meta-knowledge we expect and show the different behavior between MLN and CE and MSE. In addition, we test MLN on generated non-linear classification learning tasks with a three-layer perceptron. In actual tasks, we consider two widely used classification tasks (Mnist[19] and Cifar10 [20]) and apply well-known deep networks (LeNet[21], ResNet18[22], VGG16[23]) to solve them. Four groups of classification learning tasks are tested, including LeNet-Mnist, LeNet-Cifar10, ResNet18-Cifar10, and VGG16-Cifar10. For each task, if there is no special explanation, the reported results are the average in ten tests. We compare the performance of MLN with those of two main losses, CE and MSE.

The hyperparameters in the experiments are set up as follows. For meta-training, the optimizer of classifiers is stochastic gradient descent (SGD) with a batch size of 500 and the learning rate of 0.1. The maximum time of learning iteration is 1,000. In the ES algorithm for searching MLN, $\mu = 25$, $\lambda = 25$, each MLN is initialized with Xavier, and the initial mutation strength $\sigma$ is set to 0.05. When producing children, we change $\sigma$ of children according to Eq.5.

$$\sigma^*_i(j) = \sigma_i(j) \times e^{(\tau_0 \times N_j(0,1) + \tau_1 \times N_j(0,1))}, j = 1, \ldots, N_w \quad (5)$$

where $\tau_0 = \frac{1}{\sqrt{2 \times \sqrt{N_w}}} = 0.034537$ and $\tau_1 = \frac{1}{\sqrt{2 \times N_w}} = 0.001687$. $\sigma_i(j)$ represents the j-th mutation strength of i-th child, $N_w$ is the length of gene, $N_j(0,1)$ represents a standard normal distribution. For the meta-testing of MLN, the optimizer of classifier is changed to Adam for testing whether the learning of MLN is robust to the optimizer. In generated tasks, learning rate is 0.001 with a batch size of 500, for 1,000 steps. In LeNet-Mnist, LeNet-Cifar10, ResNet18-Cifar10, learning rate is 0.001 with a batch size of 500, for 100 epochs. And in VGG16-Cifar10, training uses Adam with a batch size of 64, for 200 epochs, since it doesn't converge steadily under the insufficient iterations. The learning rate for MLN is 0.0001, and the learning rates for CE and MSE are default values of Adam.

### 4.2 Experimental Results

### 4.1.1 For generated meta-testing tasks

First, to show what the resultant MLN is, we draw the function curve of our MLN and compare it with that of CE in Fig. 3. It illustrates that these two losses give different discrimination to the optimal classifier. For our MLN, the lowest point of loss corresponds to the prediction of 0.82 for the correct category. In contrast, the prediction corresponding to the lowest point of CE is 1. The MLN learns a new margin between two categories, narrower than the traditional 0-1 margin. It is the potential to produce a better generalization. Furthermore, the range of function values of our MLN is (2, 3.4), which is much smaller than that of CE. Compared with CE, the gradients of MLN are more uniform in the range of function values, and its gradient increases rapidly when departing from the lowest point. These two features are advantageous to gradient-based optimization.

Second, to further clarify the meta-knowledge learned in MLN, we show the changes of three losses (MLN, CE, and MSE) and meta-loss (Eq. 1) in the process of optimizing classifiers in Fig. 4. We can see that 1) all three loss functions reduce the meta-loss gradually in the early stage of classifier learning; 2) but in the later stage of classifier learning, CE and MSE show the behavior opposite from the MLN. The meta-loss starts to increase at 30 iterations; while MLN continues to optimize the classifier to reduce the meta-loss; and 3) although the three losses can decline in the process of classifier optimization, only the changing of MLN is consistent with that of meta-loss. The above results demonstrate that our meta-learning approach can successfully lead to the expected MLN that has a different learning nature from CE and MSE.

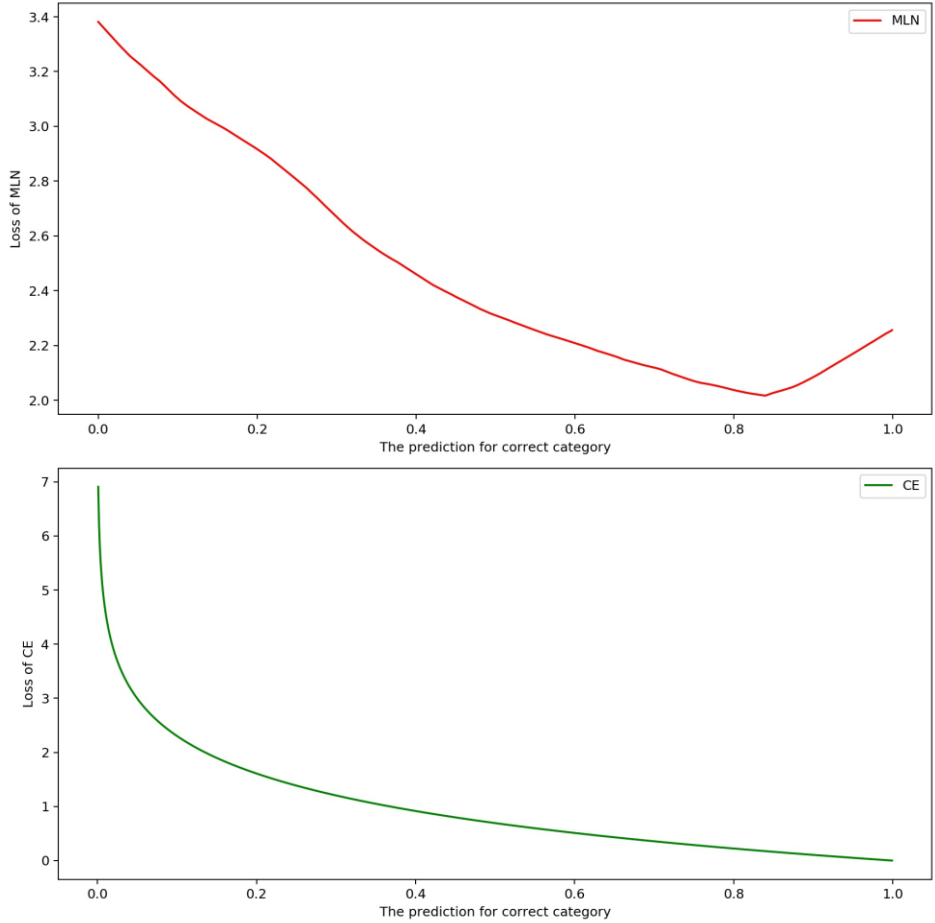

Figure 3. The function curves of MLN and CE.

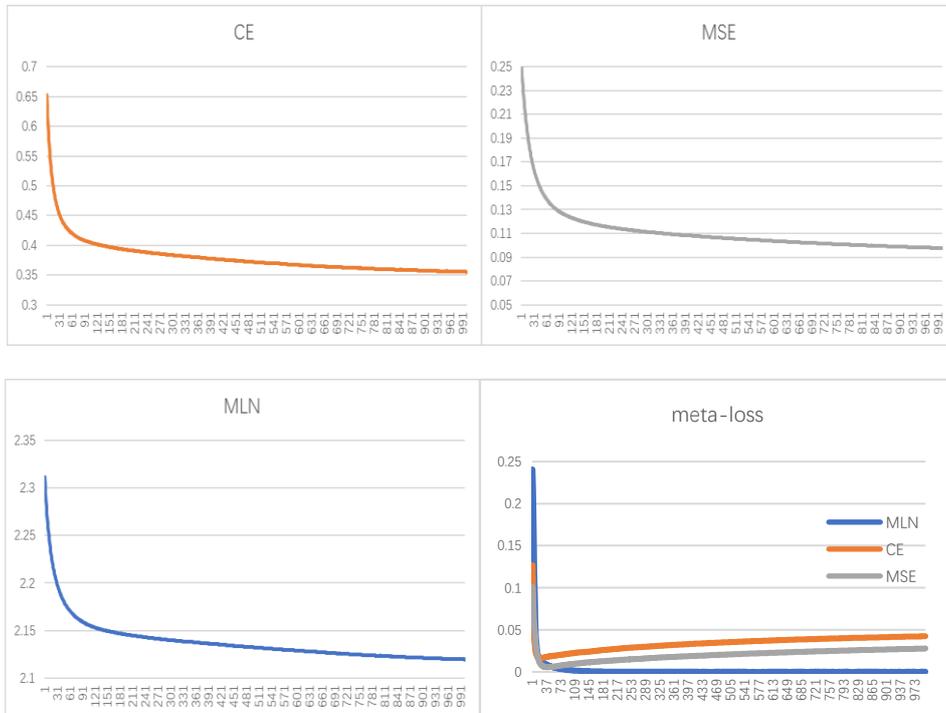

Figure 4. The changing of three losses and meta-loss in the learning of single-layer perception. (a) for CE; (b) for

MSE; (c) for MLN; (d) for meta-loss

Third, to answer the question of whether the expected MLN is also effective from the view of classification, we compare the convergence of MLN, CE, and MSE and the accuracy of the optimized classifier on the validation data on single-layer perceptron learning tasks and three-layer perceptron learning tasks, respectively. We record the changing of accuracy on validation data along with the learning process as shown in Fig.5. The corresponding results are shown in Fig. 5. In single-layer perceptron learning tasks, we can see that 1) the convergence speed of MLN is significantly faster than CE and MSE. The classifier optimized by MLN has been convergent after 500 iterations, while the CE or MSE needs 600~700 iterations to reach the same level; and 2) the finally resultant classifier from our MLN is more accurate than those from CE and MSE. The validation accuracies for MLN, CE, and MSE are 99.21%, 96.35%, and 98.12% on linear classification tasks, respectively. In three-layer perceptron learning tasks, the validation accuracies for MLN, CE, and MSE are 98.35%, 98.18%, and 98.17%, respectively.

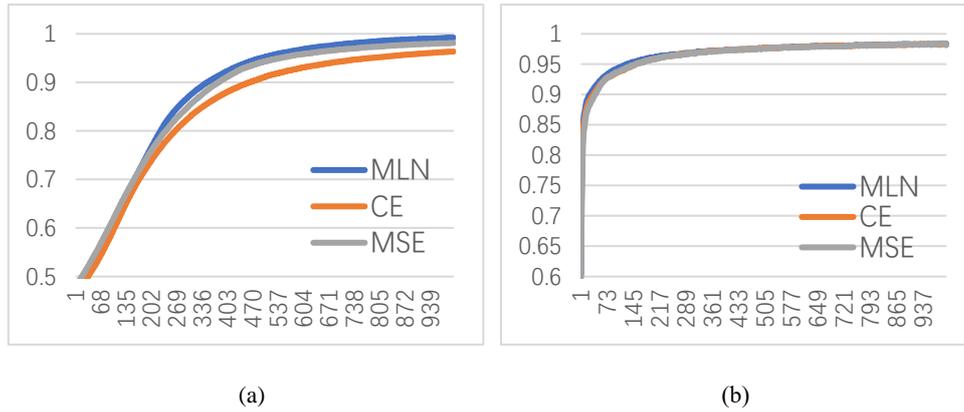

(a)          (b)

Figure 5. The validation accuracy from three losses: (a) for single-layer perception, (b) for three-layer perception.

4.2.2 For actual learning tasks

In this section, we test the performance of learned MLN on four classification learning tasks described above, i.e., LeNet-Mnist, LeNet-Cifar10, ResNet18-Cifar10, and VGG16-Cifar10. Mnist contains 60,000 training samples and 10,000 testing samples, divided into 10 classes. Cifar10 includes 50,000 training samples and 10,000 testing samples, divided into 10 classes as well.

1. Testing accuracy

The average testing accuracies in ten times of tests for each learning task are listed in Table 1. MLN is observed to outperform CE and MSE across all learning tasks. These results present an interesting conclusion that a loss function that is learned on randomly generated simplest linear learning tasks can be transferred to much more complicated learning tasks with satisfactory performance improvements.

Table 1. testing accuracies from three losses on actual learning tasks

| Dataset | classifier | MLN | CE | MSE |
|---|---|---|---|---|
| Mnist | LeNet | **99.37**% | 99.32% | 99.31% |
| Cifar10 | LeNet | **80.17**% | 80.02% | 79.82% |
| | ResNet18 | **92.36**% | 92.31% | 91.95% |
| | VGG16 | **92.41**% | 92.00% | 89.55% |

2. Generalization ability

Fig.9 presents the changing of average training loss and test loss along with the learning iterations for CE, MSE, and MLN, respectively, in four groups of actual learning tasks. MLN is found to effectively avoid overfitting of classifiers in all learning tasks of Cifar10. In the process of classifier optimization, CE and MSE have different degrees of overfitting. In the tasks of cifar10, the test loss shows the opposite trend to the train loss. But for MLN, it never happend.

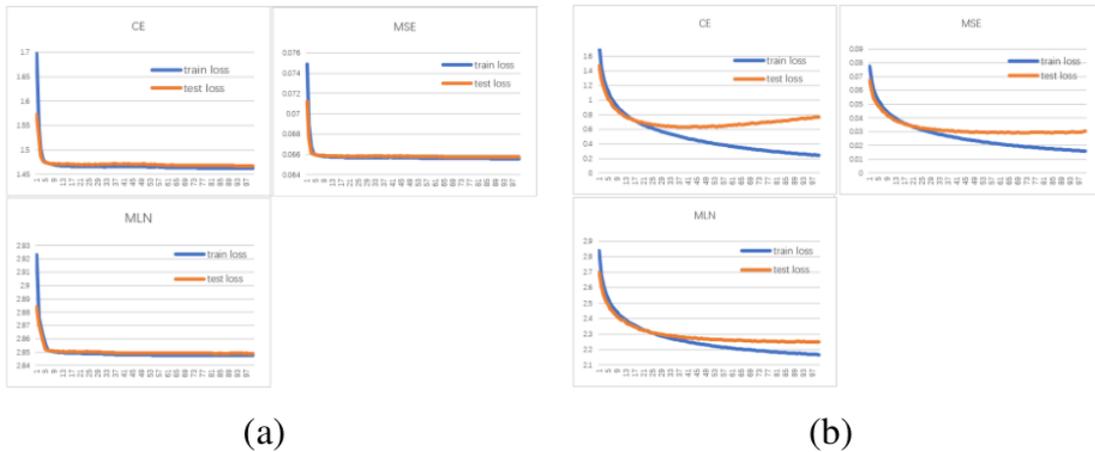

(a)          (b)

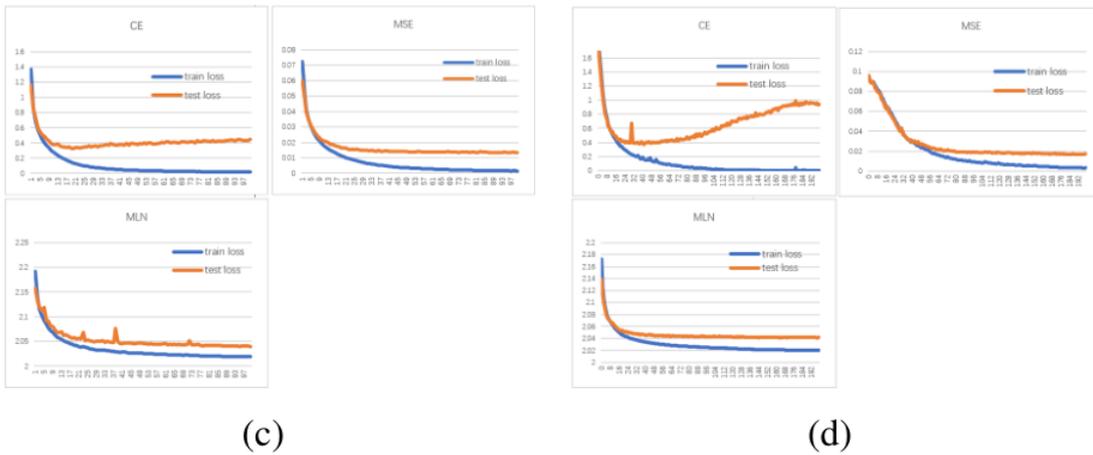

(c)          (d)

## 5. Conclusions

This paper has proposed a meta-learning approach based on randomly generated learning tasks to train a parametric loss for classification learning. The resultant loss, called MLN for short, has been evolved on generated simple tasks of linear classifier learning to stably outperform the CE and MSE on much more complicated and actual classification tasks. We have the following interesting observations: 1) it is possible and promising to conduct effective meta-learning based on randomly generated learning tasks.; 2) the difference between the learned model and the known ground-truth models is valuable for defining the satisfactory meta-learning objective.

In the following research, we will try to extend the input of our MLN from a single data point to a batch of data points for utilizing the relationship between data points to further improve the MLN's performance.

**Reference**

[1]   FINN C, ABBEEL P, LEVINE S. Model-agnostic meta-learning for fast adaptation of deep networks; proceedings of the International Conference on Machine Learning, F, 2017 [C]. PMLR.
[2]   ANDRYCHOWICZ M, DENIL M, GOMEZ S, et al. Learning to learn by gradient descent by gradient descent [J]. 2016,
[3]   GONZALEZ S, MIIKKULAINEN R. Improved training speed, accuracy, and data utilization through loss function optimization; proceedings of the 2020 IEEE Congress on Evolutionary Computation (CEC), F, 2020 [C]. IEEE.
[4]   LEE Y, CHOI S. Gradient-based meta-learning with learned layerwise metric and subspace; proceedings of the International Conference on Machine Learning, F, 2018 [C]. PMLR.
[5]   GRANT E, FINN C, LEVINE S, et al. Recasting gradient-based meta-learning as hierarchical bayes [J]. 2018,
[6]   RAVI S, LAROCHELLE H. Optimization as a model for few-shot learning [J]. 2016,
[7]   MUNKHDALAI T, YU H. Meta networks; proceedings of the International Conference on Machine Learning, F, 2017 [C]. PMLR.
[8]   RAJASEGARAN J, KHAN S, HAYAT M, et al. itaml: An incremental task-agnostic meta-learning approach; proceedings of the Proceedings of the IEEE/CVF Conference on Computer Vision and Pattern Recognition, F, 2020 [C].
[9]   BRONSKILL J, GORDON J, REQUEIMA J, et al. Tasknorm: Rethinking batch normalization for meta-learning; proceedings of the International Conference on Machine Learning, F, 2020 [C]. PMLR.
[10] VINYALS O, BLUNDELL C, LILLICRAP T, et al. Matching networks for one shot learning [J]. 2016,
[11] SNELL J, SWERSKY K, ZEMEL R S J A P A. Prototypical networks for few-shot learning [J]. 2017,
[12] SUNG F, YANG Y, ZHANG L, et al. Learning to compare: Relation network for few-shot learning; proceedings of the Proceedings of the IEEE conference on computer vision and pattern recognition, F, 2018 [C].
[13] BECHTLE S, MOLCHANOV A, CHEBOTAR Y, et al. Meta-learning via learned loss [J]. 2019,
[14] HOUTHOOFT R, CHEN R Y, ISOLA P, et al. Evolved policy gradients; proceedings of the

Proceedings of the 32nd International Conference on Neural Information Processing Systems, F, 2018 [C].

[15] WU L, TIAN F, XIA Y, et al. Learning to teach with dynamic loss functions [J]. 2018,

[16] ZHENG H, YANG Z, LIU W, et al. Improving deep neural networks using softplus units; proceedings of the 2015 International Joint Conference on Neural Networks (IJCNN), F, 2015 [C]. IEEE.

[17] GLOROT X, BENGIO Y. Understanding the difficulty of training deep feedforward neural networks; proceedings of the Proceedings of the thirteenth international conference on artificial intelligence and statistics, F, 2010 [C]. JMLR Workshop and Conference Proceedings.

[18] SCHWEFEL H-P J B U V, BASEL UND STUTTGART. Numerische Optimierung yon Computer-Modellen mittels der Evolutions-strategic [J]. 1977,

[19] LECUN Y J H Y L C E M. The MNIST database of handwritten digits [J]. 1998,

[20] KRIZHEVSKY A, HINTON G. Learning multiple layers of features from tiny images [J]. 2009,

[21] LECUN Y, BOTTOU L, BENGIO Y, et al. Gradient-based learning applied to document recognition [J]. 1998, 86(11): 2278-324.

[22] HE K, ZHANG X, REN S, et al. Deep residual learning for image recognition; proceedings of the Proceedings of the IEEE conference on computer vision and pattern recognition, F, 2016 [C].

[23] SIMONYAN K, ZISSERMAN A J A P A. Very deep convolutional networks for large-scale image recognition [J]. 2014,